\documentclass[letterpaper, 10 pt, conference]{ieeeconf}  

\IEEEoverridecommandlockouts                              

\usepackage{graphics} 

\usepackage{times} 
\usepackage{amsmath} 
\usepackage{amssymb}  
\usepackage{hyperref}
\usepackage[normalem]{ulem}
\useunder{\uline}{\ul}{}
\usepackage{multirow}
\usepackage{hyperref}

\usepackage{graphicx} 

\usepackage[caption=false, font=footnotesize]{subfig}

\usepackage{pifont}
\newcommand{\cmark}{\ding{51}}%
\newcommand{\xmark}{\ding{55}}%

\usepackage[dvipsnames]{xcolor}
\usepackage{xspace}

\newcommand{\myparagraph}[1]{\vspace{8pt}\noindent\textbf{#1}}

\newcommand{\challenge}{RGB-D Triathlon}
\newcommand{\linearscore}{{Locally Linear Score}}

\def\eg{\textit{e.g.}\xspace}
\def\ie{\textit{i.e.}\xspace}

\def\etal{\textit{et al.}\xspace}

\newcommand{\squeezeup}{\vspace{-3.5mm}}

\title{\LARGE \bf
The RGB-D Triathlon: 
Towards Agile Visual Toolboxes for Robots
}

\author{Fabio Cermelli$^{1}$, Massimiliano Mancini$^{2,3}$, Elisa Ricci$^{3,4}$ and Barbara Caputo$^{1,5}$
\thanks{This work was partially supported by the ERC project RoboExNovo (B.C.).}
\thanks{F. Cermelli and B. Caputo are with Politecnico di Torino, Turin, Italy. {\tt\small s236363@studenti.polito.it, barbara.caputo@polito.it}}%
\thanks{$^{2}$M. Mancini is with Sapienza University of Rome, Rome, Italy. {\tt\small mancini@diag.uniroma1.it}}
\thanks{$^{3}$M. Mancini and E. Ricci are with Fondazione Bruno Kessler, Trento, Italy. {\tt\small eliricci@fbk.eu}}%
\thanks{$^{4}$E. Ricci is with University of Trento, Trento, Italy.}
\thanks{$^{5}$B. Caputo is with Italian Institute of Technology, Milan, Italy.}
}

\begin{document}

\maketitle
\thispagestyle{empty}
\pagestyle{empty}

\begin{abstract}

Deep networks have brought significant advances in robot perception, enabling to improve the capabilities of robots in several visual tasks, ranging from object detection and recognition to pose estimation, semantic scene segmentation and many others. Still, most approaches typically address visual tasks in isolation, resulting in overspecialized models which achieve strong performances in specific applications but work poorly in other (often related) tasks. This is clearly sub-optimal for a robot which is often required to perform simultaneously multiple visual recognition tasks in order to properly act and interact with the environment. This problem is exacerbated by the limited computational and memory resources typically available onboard to a robotic platform. The problem of learning flexible models which can handle multiple tasks in a lightweight manner has recently gained attention in the computer vision community and benchmarks supporting this research have been proposed. In this work we study this problem in the robot vision context, proposing a new benchmark, the RGB-D Triathlon, and evaluating state of the art algorithms in this novel challenging scenario.
We also define a new evaluation protocol, better suited to the robot vision setting. Results shed light on the strengths and weaknesses of existing approaches and on open issues, suggesting directions for future research.

\end{abstract}

\section{INTRODUCTION}
Recent years have witnessed great advances in computer and robot vision thanks to deep networks \cite{krizhevsky2012imagenet}. 
Deep models are used in many applications in robot vision, ranging from egomotion estimation \cite{costante2016exploring} to depth prediction \cite{jafari2017analyzing,mancini2017toward},
object grasping \cite{johns2016deep, levine2018learning}
semantic segmentation \cite{schwarz2018rgb,oliveira2018efficient}, etc.

\begin{figure}[t]
    \centering
    \includegraphics[width=0.8\linewidth]{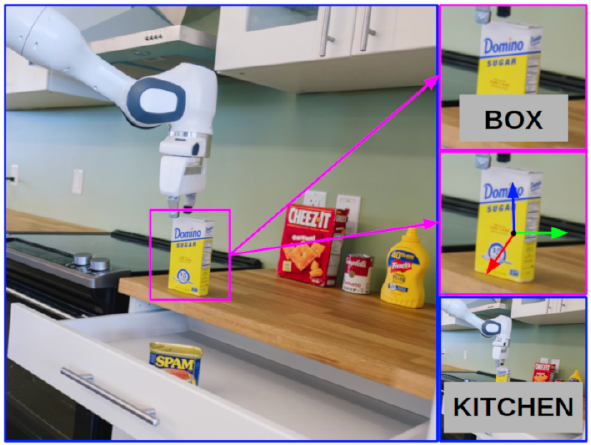}
    \caption{The ability of a robot to perform multiple tasks at the same time is crucial. For example, to manipulate correctly the object in the scene above, a robot should understand the class of the object, its orientation and the overall context where it operates. Image courtesy of NVIDIA Robotics Research Lab, Seattle (\url{news.developer.nvidia.com/nvidia-opens-robotics-research-lab-in-seattle/}).}
    \label{fig:teaser}
    \squeezeup
    \squeezeup
\end{figure}

A common procedure 
for addressing a specific visual recognition problem consists in fine-tuning an existing pretrained model on a given, problem-specific dataset.  
While this strategy leads to excellent performance on the specific problem and setting, it ignores two key aspects of robot vision. The first is that robots need visual abilities for several tasks. For instance, to complete a simple carrying task a robot needs to localize itself, detect and recognize the objects in front of it and estimate their poses (see Fig. \ref{fig:teaser}). Clearly, having an overspecialized network for each specific task would scale poorly. Second, the multiple tasks that a robot is required to solve are often closely related (see Fig. \ref{fig:teaser}). Hence, addressing the tasks jointly would probably lead to an increased accuracy as well as to an improved computational efficiency with respect to training task-specific networks.

To deal with this issue, over the years Multi-
Task Learning (MTL) \cite{Caruana1997} approaches have been developed. 
MTL refers to jointly learn a set of classification/regression models, each associated to a specific task, by leveraging information about task relatedness, \eg sharing models' parameters. By reducing the number of parameters, MTL 
also decreases the time needed for model training and inference.
MTL has been successfully applied in robotics: 
Teichmann \etal \cite{teichmann2018multinet} proposed an architecture  performing jointly object detection, classification, and semantic segmentation. Similarly, in \cite{rahmatizadeh2018vision} Rahmatizadeh \etal described a technique to train a controller to perform several complex picking and placing tasks.

MTL assumes that the data for all the tasks are available during training. In robotics this assumption is often unrealistic and the ability to add new tasks sequentially 
is crucial 
\cite{mitchell2018never, thrun1998lifelong}. However, a well-known problem of  sequential learning is that, while learning how to perform a new task, an algorithm typically forgets about previous tasks. This 
\textit{catastrophic forgetting} \cite{catforg_1, catforg_2} must be kept into account while developing visual recognition models. 

\begin{figure*} 
    \centering
  \subfloat{%
       \includegraphics[width=0.47\linewidth]{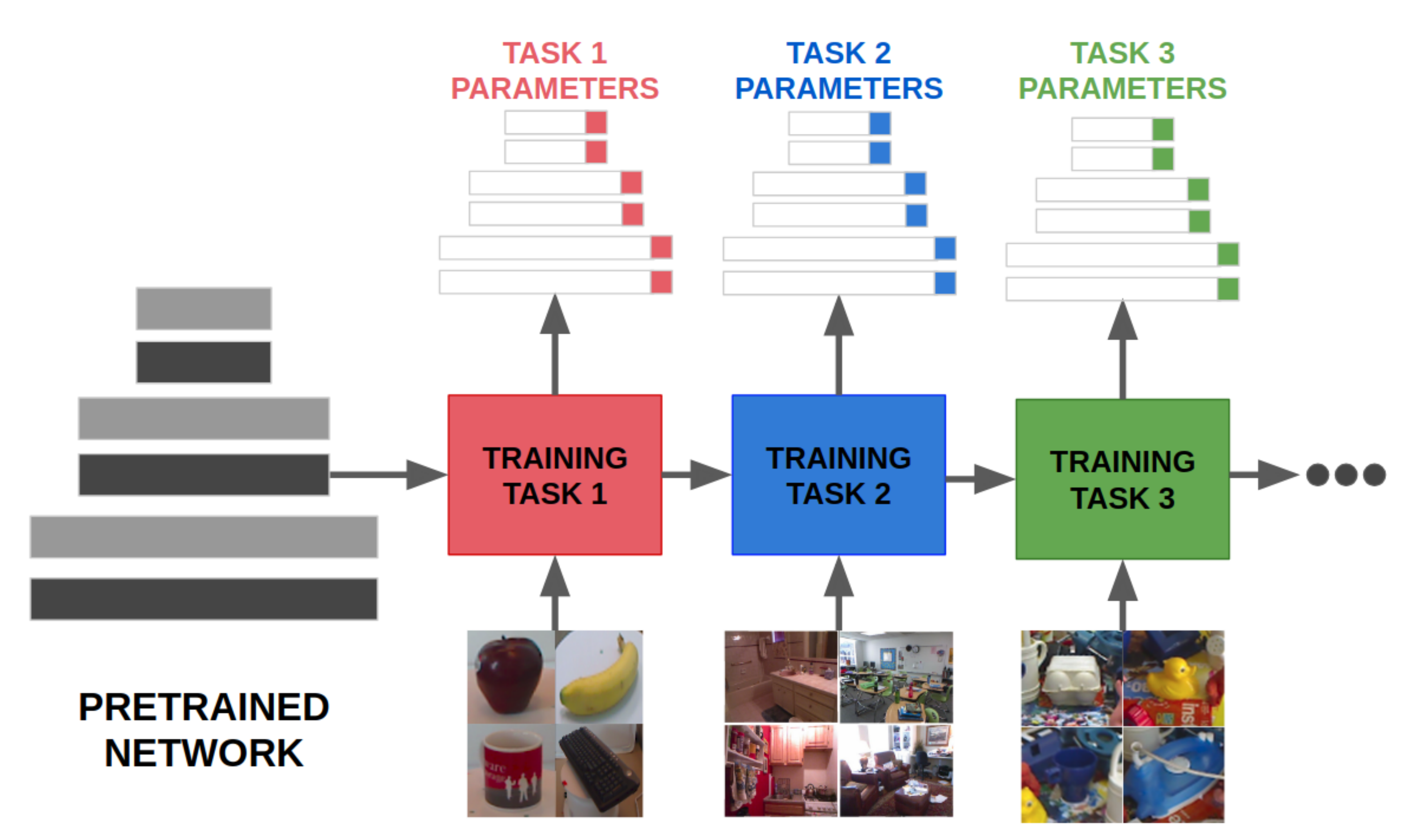}}
    \hfill
  \subfloat{%
        \includegraphics[width=0.47\linewidth]{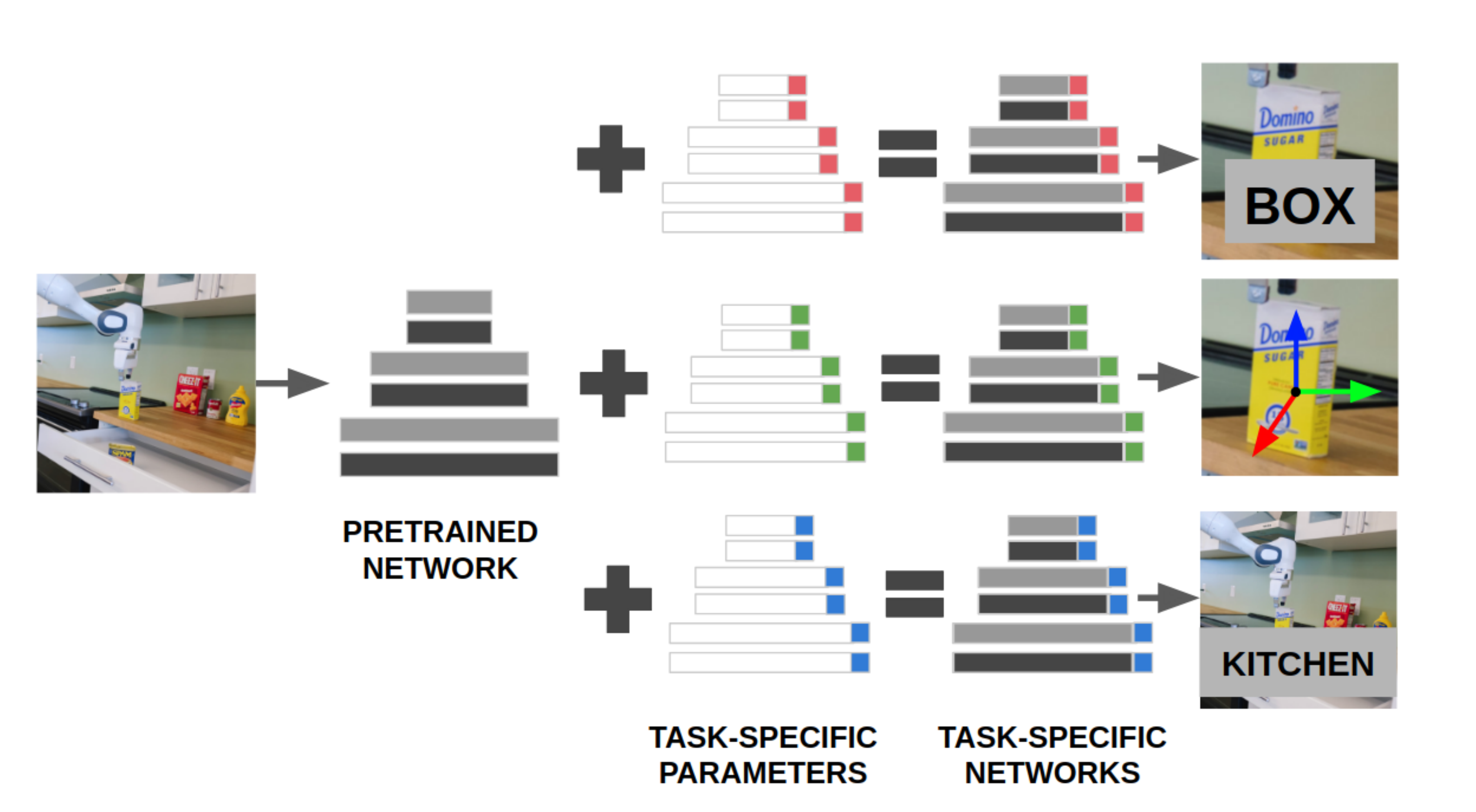}}
  \caption{The Sequential Multi-Task learning problem. During training (left) a sequence of tasks are presented to the network, one by one. For each task, a set of task-specific parameters are learned (colored blocks), freezing the shared ones (grey blocks). {During inference (right) 
  multiple tasks can be solved by combining the pretrained network with the task-specific parameters, thus keeping 
  low the overhead in terms of memory requirements.
  }}
  \label{fig:SMTL}
  \squeezeup
\end{figure*}

Sequential Multi-Task Learning (SMTL)  algorithms \cite{rebuffi_1, rebuffi_2, packnet, piggyback, quantized} automatically address the catastrophic forgetting problem by considering a common network backbone and introducing a small set of task-specific network parameters. In practice, during training, for each new task these methods instantiate and learn few task-specific parameters (see Fig. \ref{fig:SMTL}),
while the other network's weights are kept fixed. While interesting and effective, 
these approaches have been considered only in computer vision and it is not clear if they can also be used in a 
robotic setting where 
the computational/memory requirements are especially relevant. 

This paper aims at studying SMTL in the robot vision context, presenting the first 
SMTL benchmark for robot vision while testing state of the art algorithms on this setting. Moreover, to take into account the peculiarities of the robotics scenario, we propose a new evaluation protocol which allows to compare different SMTL algorithms considering not only their recognition performances but also their memory requirements and their ability to deal with multiple/different input modalities (\ie RGB-D).




\myparagraph{Contributions.}
{To summarize, the contributions of this work are three-fold.
\begin{itemize}
    \item We propose the first benchmark for sequential multi-task learning in  robotics, the RGB-D Triathlon. 
    It considers different input modalities and three fundamental tasks: object recognition, 
    pose estimation 
    and scene classification. 
    On the considered dataset, we evaluate several state of the art SMTL methods: the serial \cite{rebuffi_1} and parallel \cite{rebuffi_2} residual adapters, Piggyback \cite{piggyback} and Binarized Affine Transfom (BAT) \cite{quantized}.
    \item We propose a new metric for the SMTL task which considers both model accuracy and the memory/computational requirements. 
    \item We release a toolbox enabling researchers to develop their own method for SMTL and compare it with baseline methods. Our code 
    can be found at \url{https://github.com/fcdl94/RobotChallenge}.
\end{itemize}
 
}

\section{RELATED WORKS}
Our work is related to many previous studies in the area of visual learning, \eg those addressing the problems of incremental and multi-task learning. 
In the following we first review recent benchmarks
proposed in computer and robot vision involving multiple tasks. Then we describe the relation between our work and previous studies on incremental and multi-task learning. We consider only deep neural models, due to their clear advantages in performance compared to shallow models.

\myparagraph{Learning from Multiple Tasks/Domains: Benchmarks.}
Benchmarks are fundamental tools to advance research in visual recognition and robot perception. 
For instance, in robot vision, datasets as SeqSLAM \cite{milford2012seqslam}, RGB-D SLAM \cite{sturm2012benchmark}, allowed the development of methods for addressing loop-closing, SLAM \cite{mur2015orb} and semantic mapping \cite{sunderhauf2017meaningful}.
%
In computer vision, datasets and challenges as ImageNet \cite{imagenet}, Common Objects in Context (COCO) \cite{lin2014microsoft} and Places \cite{zhou2018places} have brought significant progresses, enabling to learn deep architectures which are not only effective for a single categorization problem 
but that can also serve as general purpose models to address other recognition tasks 
\cite{krizhevsky2012imagenet}. 

Recently, the idea of learning universal representations \cite{bilen2017universal} and task-agnostic deep models, 
able to perform well over multiple tasks and/or domains, 
has attracted attention \cite{rebuffi_1,quantized}, fostering the creation of new datasets and challenges. 
A notable example is the Robust Vision Challenge (http://www.robustvision.net), which aims to facilitate the development of robust models, \ie models which must perform well on a specific problem (\eg depth estimation, semantic segmentation) but on different settings (\eg environments, sensors). 
Another interesting benchmark is the Visual Domain Decathlon \cite{rebuffi_1}, which promotes 
the developments of SMTL models considering ten different categorization problems. Drawing inspiration from these initiatives in the computer vision community, in this paper we propose a novel benchmark to stimulate research in robot perception. In particular, with respect to the aforementioned SMTL benchmarks, we consider a setting where i) the tasks to be addressed are different (namely, pose estimation, object recognition and scene classification); ii) the evaluation protocol is newly designed, in order to take into account not only the per-task performances but also
the computational complexity of the model; 
iii) different input modalities are considered (\ie RGB, Depth and RGB-D).

\myparagraph{Sequential Multi-Task Learning. } Given a pretrained model and a set of tasks whose data are available at \textit{different} times, the goal of Sequential Multi-Task Learning \cite{quantized} is to learn a network able to address all tasks while keeping low the overhead in terms of required parameters. In computer vision, this task has been tackled by methods addressing the aforementioned Visual Domain Decathlon challenge~\cite{rebuffi_1,rebuffi_2,piggyback,quantized}. These models differ on how they extend a pretrained model to new tasks, considering for instance adding task-specific network modules~\cite{rebuffi_1,rebuffi_2} or changing the network parameters by means of binary masks \cite{piggyback,quantized}. In robot perception, this task has not been addressed yet, thus we will take these models as baseline approaches.

\myparagraph{Multi-Task and Incremental Learning. }
The formulation of our task is strictly related to 
MTL and incremental learning. Similar to ours, the goal of MTL is to train a model on multiple tasks. 
Different works addressed MTL in various contexts of robot perception, ranging from reinforcement learning \cite{devin2017learning}, learning from demonstrations \cite{rahmatizadeh2018vision}, grasping \cite{fang2018multi} and scene understanding \cite{teichmann2018multinet}. {Differently from standard MTL settings, we focus on the sequential setting, where we do not have access to the data of all the tasks beforehand but we receive them one after the other, sequentially.} 

SMTL is also closely related to incremental learning \cite{thrun1998lifelong}. Our aim is also to progressively add knowledge to a model without forgetting the previously learned concepts, overcoming the problem of catastrophic forgetting~\cite{french1999catastrophic}. Still, differently from the incremental and the incremental class learning problems \cite{lomonaco2017core50,valipour2017incremental,camoriano2017incremental,turkoglu2018incremental}, we want to add tasks to the network and not consider new categories, thus we need a separate output space for each new task. 

\section{An SMTL benchmark for Robot Perception
}
\label{sec:challenge}

\begin{figure*} 
    \centering
  \subfloat[RGB-D Object dataset \cite{rod}]{%
       \includegraphics[width=0.3\linewidth]{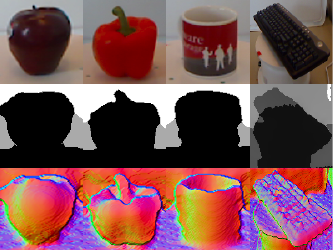}}
    \hfill
  \subfloat[NYU Depth V2 \cite{nyu}]{%
        \includegraphics[width=0.3\linewidth]{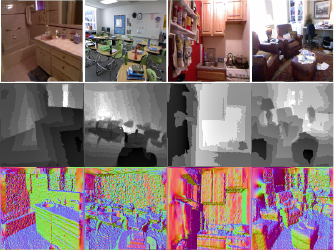}}
    \hfill
  \subfloat[{LineMOD} \cite{linemod}]{%
        \includegraphics[width=0.3\linewidth]{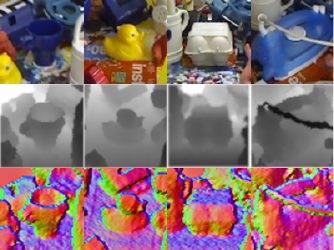}}
    \hfill
  \caption{Images taken from the datasets included in the benchmark. For each image, the top row shows the RGB version, the middle row the raw depth, and the bottom row the depth images colored with the Surface method \cite{aakerberg2017depth}.}
  \label{fig:datasets}
  \squeezeup
\end{figure*}

As stated above, the aim of this work is to foster research in developing robot systems which are able to sequentially learn to perform several visual tasks by applying SMTL methodologies. This is crucial for many reasons. First, the memory resources typically available in a robotic platform are limited, thus it is practically unfeasible to store a new deep network for each novel visual task a robot is asked to solve. Second, we may be interested in extending the visual capabilities of an existing robotic system, in order to solve new tasks not considered at the initial stage of deployment. 
In a nutshell, SMTL techniques aim to learn a set of classification/regression models for multiple tasks sequentially. {The challenge is to learn new task-specific models while keeping the number of task-specific parameters as low as possible.}

{Formally, the SMTL task is defined as follows. Suppose we have a pretrained deep model, with parameters $\Theta_0$ and a set $\mathcal{T}=\{\mathcal{T}_1,\cdots,\mathcal{T}_N\}$ of $N$ tasks. For each task we have an input space $\mathcal{X}\subset \Re^{H\times W \times C}$, with $C$ depending on the input modality (\ie 1 for depth maps, 3 for RGB images and 4 for RGB-D), a feature space $\mathcal{Z}_t$ and an output space $\mathcal{Y}_t$. For simplicity, let us define as $\phi_{\Theta_t}:\mathcal{X}\rightarrow\mathcal{Z}_t$ a task-specific mapping from the input space to the feature space $\mathcal{Z}_t$. The mapping is parametrized by the set $\Theta_t=\{\theta_0, \ \theta_t\}$ which comprises the shared ($\theta_0\in\Theta_0$) and task-specific ($\theta_t$) parameters. Moreover, let us denote as $\psi_{\Omega_t}:\mathcal{Z}_t\rightarrow\mathcal{Y}_t$ the function mapping the feature space to the task-specific output space, parametrized by $\Omega_t$. In our case, $\Omega_t$ are just the weights of the output layer of our neural network. 
Finally, let us define the mapping from the input to the output space $\Phi_{t}:\mathcal{X}\rightarrow\mathcal{Y}_t$, where $\Phi_t=\psi_{\Omega_t} \circ \phi_{\Theta_t}$. 
We also denote as $\rho_t=|\theta_t|$, the memory size (in bits) required to store the task-specific parameters $\theta_t \in \Theta_t$ and as $\alpha_t$ a task-specific performance measure (\eg classification accuracy). 
The goal of an SMTL algorithm is to learn a set of task-specific mappings $\Phi_t$ while i) maximizing $\alpha_t$ and ii) keeping $\rho_t$ as low as possible for any $\mathcal{T}_t \in \mathcal{T}$.

To stimulate the research community in robot vision on SMTL, in this paper we propose a dataset consisting of three different tasks and introduce an evaluation metric suitable to the STML setting. The dataset is a collection of three datasets commonly used in the robotics community, but in this paper we propose their joint adoption for studying the novel problem of SMTL.}

{In the following subsections, we present the dataset and the tasks we proposed. Then, we describe the novel evaluation protocol we introduced for SMTL in robotics.}

\subsection{The \challenge \hspace{2pt}Dataset}
\label{tasks}
The proposed dataset considers three fundamental tasks: object recognition, pose estimation and scene classification. In choosing the tasks and benchmarks we follow three principles, namely i) the importance of the perception task for robotics; ii) the impact that the dataset has had in the past within the community; iii) the possibility of defining a standard training/testing protocol. Differently from challenges in computer vision \cite{rebuffi_1}, we propose three different settings: i) use only RGB images, ii) use only depth images, iii) use both RGB and depth images. Each setting is independent from the others, such as to permit the researchers to work only on the setting which is more relevant to their application. 

\myparagraph{Task 1. Object Recognition.}
Object recognition is of utmost importance in robot perception. 
This task consists in assigning to an input image depicting an object the corresponding semantic class label. Object recognition algorithms are fundamental tools for robots because i) knowing the category of an object allows reasoning on how the object can be manipulated and ii) they enable more complex tasks such as object detection. 

We consider the popular RGB-D Object Dataset (ROD) \cite{rod}. It contains 300 common household objects organized in 51 categories. The dataset was recorded using a Kinect style 3D camera that gathers 30 synchronized and aligned $ 640 \times 480$ RGB and depth images per second. Each object was placed on a turntable during the recording and it was captured during a whole rotation. For each object, there are three video sequences, corresponding to the camera mounted at a different height (so that the object is seen from different angles, \ie approximately 30, 45, and 60 degrees relative to the horizon).
The performance of a model on this task is evaluated as the accuracy on the test samples. We use the same evaluation protocol of \cite{rod} subsampling the dataset by taking every fifth frame, resulting in 41,877 RGB-D images for training and evaluation. They defined 10 pre-defined training and test splits for cross-validation, and in each split, one random object instance from each class is left out from the training set and used for testing. This results in roughly 35,000 training images and 7,000 test images in each split. At test time, the classifier has to assign the correct label to a previously unseen object instance from each of the 51 classes. For training and test we use the first split among the ten available.


\myparagraph{Task 2. 3D Pose Estimation.}
Pose estimation refers to the task of predicting the pose of an object with respect to the viewer's camera. This task is important in robotics, \eg due to the information it provides for grasping and manipulating an object \cite{johns2016deep}. The pose estimation task can be decomposed in two sub-tasks: the localization of the object in the image and the estimation of the rotation matrix between the object and the camera. Even though many works attempt to perform both sub-tasks at the same time \cite{bb8,poseCNN}, following \cite{poseregression, learningdescriptors} in this paper we assume that the object is always centered in the input image, reducing the pose estimation task only to the latter sub-task, \ie 3D pose estimation.

We consider the LineMOD dataset \cite{linemod}. We choose this dataset since it is a standard benchmark in robotics and it also provides depth information, as opposed to other benchmarks \cite{xiang2014beyond}. It contains 18,000 RGB-D images of 15 different objects classes. We adopt the cropped version of the dataset that was proposed in \cite{learningdescriptors}.
In this version all the images are squared with size $64 \times 64$ pixels and they contain the objects centered in the scene. The ground truth of the pose is given in the form of $3 \times 3$ rotation matrix that maps the camera world coordinate into the camera coordinates. 

Some methods tested on this dataset consider the use of both synthetic and real images 
\cite{zakharov20173d,wohlhart2015learning}. 
In this paper, in order to have a simpler experimental setup, we follow \cite{porzi2017depth} and we consider a setting where only real images are used for training and testing the models. In particular, we use $80\%$ of the images for training and 20$\%$ for testing. 

%

For evaluating the performances of a model in the pose estimation task we consider both the prediction of the pose of the object and its semantic class. In particular, we consider a pose to be correctly estimated if the predicted object class is correct and the geodesic error of the predicted rotation is less then 20 degrees. 

\myparagraph{Task 3. Scene Classification.}
This task consists of assigning to an image a label that indicates the place where the picture was taken. Semantically localizing a robot provides relevant information on how the robot should interact with the environment. 

We use the popular NYU-Depth V2 dataset \cite{nyu}, employed by many previous works \cite{song2017depth,song2017combining,du2018depth}. The dataset contains 1,449 pairs of synchronized RGB and depth images, gathered from a wide range of commercial and residential buildings in three different US cities, comprising 464 different indoor scenes across 27 scene classes. Anyway, most of these scene classes are not well represented, thus, following  \cite{Gupta_2013_CVPR}, we reorganized the original 27 categories into 10 classes (\ie the 9 most represented categories and the rest).
The dataset is split in training and test data. Following \cite{nyu}, 795 images belong to training data and 654 to the test set. The performance of the model is evaluated as the classification accuracy on the test set.

Finally, we show sample images (RGB, depth and colorized depth~\cite{aakerberg2017depth}) for each dataset included in our benchmark in Figure \ref{fig:datasets}.

\subsection{The Evaluation Metrics}
The \challenge, as we defined it in section \ref{sec:challenge}, requires a new evaluation metric, going beyond the mere comparison of standard accuracies. To this extent, in the following we define a metric which takes into account the two properties a good SMTL algorithm should have: i) single-task performances as close as possible to those of task-specific architectures 
ii) number of task-specific parameters as low as possible.

{To define the metrics, we will consider two standard transfer learning methods: \textit{fine-tuning} and \textit{feature extractor}.
\textit{Fine-tuning} (FT) replicates the pretrained backbone architecture for every task and fine-tunes it independently: this produces a different network for each task. Since the full architecture is replicated and fine-tuned, $\rho_t=\rho_0$, where $\rho_0=|\Theta_0|$ denotes the memory size required to store the parameters of the pretrained backbone architecture, excluding the final output layer.
\textit{Feature extractor} (FE) keeps a single backbone architecture but for each task it instantiates a new output layer. The weights of the network are frozen (\ie they are not optimized) and only the task-specific output layers are learned. In this model, $\rho_t=0$ since $\theta_t=\emptyset$ and only the classifier $\psi_{\Omega_t}$ is learned. In the text we will denote as $\alpha_t^{\text{FE}}$ and $\alpha_t^{\text{FT}}$ the performance on the task $\mathcal{T}_t$ of the feature extractor and fine-tuning baseline respectively.
}

In the following, we review standard and previous evaluation metrics (\ie \cite{rebuffi_1}), highlighting their drawbacks and proposing a new metric, the \linearscore, overcoming them.



\myparagraph{Average Accuracy (AvgA). }
Obviously, the most straightforward metric one can adopt is the average accuracy per task. Assuming the task-specific performance measures (\ie accuracies) to be directly comparable (\ie normalized on the same range) 
we can compute the performance score as:
\begin{equation}
    S_{\text{AvgA}} = \frac{1}{N}\sum_{t=1}^N \alpha_t.
\end{equation}
Obviously, this choice has the drawbacks of not considering i) the complexity of each task and ii) the amount of memory required to store the task-specific parameters.

\myparagraph{Decathlon Score (DS). }
In order to jointly consider all the tasks and measuring the performance of SMTL methods, Rebuffi \etal proposed the Decathlon Score \cite{rebuffi_1}.
This metric 
favors methods that perform well on all tasks at the same time over methods which have mixed performances (\ie high accuracy on some tasks and low on the others). 
The metric is based on the definition of a minimal accuracy per task $\alpha_t^{\text{min}}$ under which the score obtained for the task $\mathcal{T}_t$ is zero. The minimal accuracy is set in~\cite{rebuffi_1} to the accuracy obtained by doubling the error on the task $\epsilon_t^{\text{FT}}=1-\alpha_t^{\text{FT}}$ of the task-specific fine-tuned model, namely:
\begin{equation}
   \alpha_t^{\text{min}}=\text{max}(0,1-2\cdot\epsilon_t^{\text{FT}})=\text{max}(0,\,2\cdot\alpha_t^{\text{FT}}-1) 
\end{equation}
The overall score is computed as follows:
\begin{gather}
	S_{\text{DS}} = \sum_{t=1}^{N}\eta_t \ \text{max} (0, \alpha_t - \alpha^{\text{min}}_t)^{\gamma_t},\\
	\eta_t = 1000\cdot( 1 - \alpha^{\text{min}}_i)^{-\gamma_i}
\end{gather}
where $\gamma_t\geq1$ {is a coefficient which rewards accuracy improvements and it has been set to $2$ in \cite{rebuffi_1}}. The parameter $\eta_t$ normalizes the score of each task in order to constrain it in the range $[0,1000]$. 
The advantage of this score is that it emphasizes the consistency of the performances across tasks, 
penalizing models which achieve very good performances but just on few tasks.

While this metric takes into account the complexity of each task by defining $\alpha^{\text{min}}_t$,  
it does not contain any term reflecting the amount of memory required for storing the task-specific parameters. 
A possibility to include this in the score is by revisiting DS as follows (RevDS):
\begin{equation}
	S_{\text{RevDS}} = \sum_{t=1}^{N}\eta_t\cdot \lambda^{-\frac{\rho_t}{\rho_0}}\cdot \ \text{max} (0, \alpha_t - \alpha^{\text{min}}_t) ^{\gamma_t}
\end{equation}
where $\lambda>1$ is a coefficient weighting the impact of the memory size required to store the task-specific parameters $\rho_t$. 
The higher is $\lambda$, the higher is the impact of the parameters. We set $\lambda=10$ in our experiments. 

This metric preserves the positive aspects of the DS while considering also the memory consumption.
However, it requires to set several parameters (\eg $\lambda$, $\gamma$) and it does not take into account the actual benefits that the task-specific parameters may bring with respect to a baseline where just the final output layer is learned. 

\begin{table}[t]
\centering
\caption{A comparison of the properties of SMTL metrics}
\label{tab:metrics}
\resizebox{\linewidth}{!}{%
\begin{tabular}{c|cccc}
\multicolumn{1}{l|}{} & \multicolumn{1}{c}{\begin{tabular}[c]{@{}c@{}}considers \\complexity\\of tasks\end{tabular}} & \multicolumn{1}{c}{\begin{tabular}[c]{@{}c@{}}penalizes\\additional\\parameters\end{tabular}} & \multicolumn{1}{c}{\begin{tabular}[c]{@{}c@{}}compares\\to feature\\extractor\end{tabular}} & \multicolumn{1}{c}{\begin{tabular}[c]{@{}c@{}}compares\\to fine\\tuning\end{tabular}} \\ \hline
Avg.A & \xmark & \xmark & \xmark & \xmark \\ \hline
DS & \cmark & \xmark & \xmark & \cmark \\ \hline
RevDS & \cmark & \cmark & \xmark & \cmark \\ \hline
LL & \cmark & \cmark & \cmark & \cmark \\ \hline
\end{tabular}}
\squeezeup
\end{table}

\myparagraph{\linearscore\hspace{2pt}(LL).}
In this work we propose a novel metric which considers both the memory requirement and the accuracy, while getting rid of the coefficients required by DS and RevDS. It is computed as follows:
\begin{equation}
\label{eq:ll}
S_{\text{LL}}=\frac{1000}{N}\sum_{t=1}^{N} R_t \cdot A_t
\end{equation}
where:
\begin{equation}
\label{eq:ll-details}
R_t = \text{max}(0,1-\frac{\rho_t}{\rho_0}) \;\;\;\;\text{and}\;\;\;\;A_t=\frac{\text{max}(0, \alpha_t-\alpha_t^{\text{FE}})}{\alpha_t^{\text{FT}}-\alpha_t^{\text{FE}}}
\end{equation}


This metric contains two terms. The first element, $R_t$, penalizes the increase of the memory size required to store the task-specific parameters. The second element, $A_t$, normalizes the single task performance by considering i) the gain obtained by introducing task-specific parameters $\alpha_t$ with respect to not introducing them (\ie $\alpha_t^{\text{FE}}$) and ii) the ratio between this gain and the one obtained by fine-tuning the full architecture (\ie $\alpha_t^{\text{FT}}$). 

From Eq.~\eqref{eq:ll} and ~\eqref{eq:ll-details}, if an SMTL model does not require any task-specific parameter (\ie $\rho_t=0$), $R_t=1$ and only the second term $A_t$ will be considered for computing the score. In the case $\rho_t>0$, $A_t$  will be linearly scaled by the ratio $\rho_t/\rho_0$. $S_{\text{LL}}=0$ if the size of the task-specific parameters is equal (or greater) to the size of the shared ones, as in the full fine-tuning case where $\rho_t=\rho_0$.

For the accuracy component $A_t$ the rationale is similar. Since we require the single-task performance $\alpha_t$ of an SMTL model to be at least better than the performance obtained by not adding any parameter (\ie the feature extractor baseline, $\alpha_t^{\text{FE}}$), we set the score to zero if $\alpha_t\leq \alpha^\text{FE}_t$. At the same time, we use the difference among $\alpha^{FE}_t$ and $\alpha^{FT}_t$ as a normalization factor to check how well a model is able to fill the performance gap existing between the task-agnostic baseline (\ie $\alpha_t^{\text{FE}}$) and the full task-specific counterpart (\ie $\alpha_t^{\text{FT}}$). The properties of this new metric and a comparison with the metrics previously defined is reported in Table \ref{tab:metrics}.

Finally, we highlight that while in the current version of the challenge $N=3$, the evaluation metrics we have defined are applicable to any number of tasks and to any backbone architecture, given the corresponding $\alpha_t^{\text{FE}}$ and $\alpha_t^{\text{FT}}$. This will allow to easily extend the benchmark in the future. Moreover, we \textit{do not} specify any fixed order for the tasks: this choice allows researchers to experiment with different sequences, with the possibility of exploring the relations among the tasks.

\section{EXPERIMENTS}
\label{sec:experiments}

In this section, we test state of the art SMTL algorithms \cite{rebuffi_1,rebuffi_2,piggyback,quantized} on our \challenge. We first present the baseline methods and describe the implementation details, then we discuss the results of our evaluation. 

\subsection{SMTL methods}
\label{sec:baselines}
{Together with the FT and FE baselines, introduced in Section~\ref{sec:challenge}, we evaluated four state of the art SMTL methods in our experiments:}
\textit{series} \cite{rebuffi_1} and \textit{parallel residual adapters} \cite{rebuffi_2}, \textit{Piggyback} \cite{piggyback}, and \textit{binarized affine transformation (BAT)} \cite{quantized}. 



In the \textit{series residual adapter} (RS)~\cite{rebuffi_1} task-specific parameters correspond to residual adapter modules added in series after the convolutional layer of each residual unit.  
In \textit{parallel residual adapter} (RP)~\cite{rebuffi_2} the task-specific modules are added in parallel to the convolutional layers of each residual block and not serially. In both cases, the residual adapter is implemented as a convolutional layer with kernel size $1 \times 1$ (plus a batch-normalization layer in \cite{rebuffi_1}). 

\textit{Piggyback} (PB)~\cite{piggyback} makes use of task-specific binary masks that are added to each convolutional layer of a backbone network. These binary masks multiply point-wise the original network weights, de facto producing new convolutional filters for the task of interest. 

Similarly to \cite{piggyback}, \textit{BAT}~\cite{quantized} uses task-specific binary masks that are paired with each convolutional layer of a backbone network. Differently from~\cite{piggyback} the masks are not applied through a point-wise multiplication but are used to perform an affine transformation of the convolutional filters, creating a new set of task-specific weights. 

\subsection{Implementation}
All the methods we evaluated require a pretrained architecture. We use the \textit{ResNet-18}~\cite{resnet} pretrained on ImageNet~\cite{imagenet}. This network has been chosen as it guarantees a good trade-off between accuracy and speed.

All the methods have been tested on three settings, corresponding to different inputs: RGB only, depth only and RGB-D. To handle the depth images we took inspiration from \cite{aakerberg2017depth}, processing the depth images using the Surface++ approach. For the single modality case, we simply consider as backbone network the ResNet-18. 
For the RGB-D setting we fuse by concatenation the features of two ResNet-18, one processing only RGB images and the other only depth maps. 
The fused features are passed through a fully connected layer that we use as the output layer, similarly to \cite{eitel2015multimodal}. 

To fairly compare all the methods, the same set of hyperparameters is used for all the methods in each task. The networks are trained using Stochastic Gradient Descent (SGD)~\cite{bottou2010large} with momentum except for Piggyback and BAT where SGD is used for the classification layer and Adam \cite{kingma2014adam} is used for the rest of the network, as suggested in \cite{piggyback, quantized}.
The networks are trained for 30 epochs in each setting, with a batch-size equal to 32 and weight decay $5\cdot10^{-5}$. 
The learning rate of the SGD optimizer is set to 0.005 for every task and method, while the Adam learning rate is adapted for each task, using $1\cdot10^{-5}$ for ROD, $5\cdot10^{-5}$ for LineMOD and $1\cdot10^{-4}$ for NYU. The learning rates are decayed by a factor of 10 after 20 epochs. 


To implement all the baselines we used the \href{pytorch.org}{PyTorch} framework. 
The code of the networks and the training procedure are publicly available\footnote{\url{https://github.com/fcdl94/RobotChallenge}}.

\begin{table}[t]
\caption{
Comparison between ResNet-18 \cite{resnet} fine-tuned on the task and state of the art methods in the RGB-D setting.}
\label{tab:sota}
\centering
\resizebox{1.\linewidth}{!}{%
\begin{tabular}{llccc}
\multicolumn{1}{c}{}                                                    & \multicolumn{1}{c}{}          & \textbf{ROD} & \textbf{LineMOD} & \textbf{NYU} \\ \hline
\multirow{3}{*}{\begin{tabular}[c]{@{}c@{}}\textbf{ROD}\end{tabular}}            & \multicolumn{1}{c|}{CNN+Fisher \cite{li2015hybrid}}                     &     93.8     & -                & -            \\  
                                                                        & \multicolumn{1}{c|}{(DE)$^2$CO \cite{carlucci2018deco}}                 &     93.6     & -                & -            \\ 
                                                                        & \multicolumn{1}{c|}{FusionNet enhanced \cite{aakerberg2017improving}}   &     93.5     & -                & -            \\ \hline
\multirow{2}{*}{\begin{tabular}[c]{@{}c@{}}\textbf{LineMOD}\end{tabular}} & \multicolumn{1}{c|}{Zakharov et al. \cite{zakharov20173d}}            & -            &     93.2\footnotemark        & -            \\  
                                                                        & \multicolumn{1}{c|}{Wohlhart et al. \cite{learningdescriptors}}   & -   &   96.2\footnotemark[\value{footnote}]    & -     \\ \hline 
\multirow{2}{*}{\begin{tabular}[c]{@{}c@{}}\textbf{NYU}\end{tabular}}    & \multicolumn{1}{c|}{Du et al. \cite{du2018depth}}                               & -            & -                &      67.5    \\  
                                                                        & \multicolumn{1}{c|}{Song et al. \cite{song2017depth}}                   & -            & -                &      66.7    \\
                                                                        \hline
Ours                                                                    & \multicolumn{1}{c|}{ResNet-18}                                          &     93.9     &       96.9       &      68.4   \\ \hline
\end{tabular}%
} \squeezeup
\end{table}
\footnotetext{The metric does not consider classification}

\subsection{Results}

\vspace{-2.5mm}
\myparagraph{Comparison with state of the art.} Before analyzing the performance of SMTL methods, we conduct a preliminary experiment to evaluate the performance of ResNet-18 with FT on the considered tasks. In Table \ref{tab:sota} we report our results on the RGB-D setting. 

For ROD \cite{rod} we only report the three best performing methods in the literature \cite{li2015hybrid, carlucci2018deco, aakerberg2017improving}. In \cite{aakerberg2017improving} an ensemble of deep models is used, extracting depth information through different colorization techniques. In \cite{carlucci2018deco} a deep architectures is considered to learn how to map depth data to three channel images. In \cite{li2015hybrid} depth data are encoded with Fisher vectors without adopting colorization approaches. 

For the LineMOD dataset \cite{linemod} we consider two state of the art approaches \cite{zakharov20173d, learningdescriptors}. In \cite{learningdescriptors} a convolutional network  is used to map the image space to a descriptor space where the pose and object classes are predicted through a nearest neighbour classifier. 
The method in \cite{zakharov20173d} builds upon \cite{learningdescriptors}, introducing a triplet loss function with a dynamic margin. These works employ a slightly different settings than ours since they use synthetic images. We report as accuracy measure a metric that consider correct a test image only if the angular error is below 20 degree.

For the NYU Depth V2 dataset \cite{nyu} we report two state of the art methods \cite{du2018depth, song2017depth}. Du \etal \cite{du2018depth} introduce a two-step training strategy 
that translates RGB images to depth ones. In \cite{song2017depth} 
Song \etal propose to learn from scratch the depth features with the help of depth patches. 

The results in Table \ref{tab:sota} clearly show a fine-tuned ResNet-18 is a competitive baseline by itself on each task, confirming the appropriateness of our choice.

\myparagraph{Results on the \challenge.}
In this subsection, we compare SMTL methods on our \challenge. We first analyze the performance of all SMTL techniques and compare it with FT and FE. Table \ref{tab:accuracy} reports the accuracy for each method, setting and task. AvgA$_\%$ denotes the average accuracy in percentage. Looking at the results for different modalities, in two out of three settings FT guarantees the best accuracy. This is somehow expected as this corresponds to use a separate network for each task. Moreover, FE corresponds to the worst performance, as considering the backbone network purely as a feature extractor is a sub-optimal strategy. Interestingly, the SMTL methods are very competitive with FT, despite they use few task-specific parameters (see also Table \ref{tab:score}). In particular, in the RGB setting BAT outperforms FT. We ascribe this behavior to the regularization effect introduced by the use of binary masks. 
Comparing different SMTL techniques, in the Depth only and RGB only settings BAT outperforms RS, RP and PB. 

\begin{table}[t]
\caption{Results in term of accuracy for each task in the \challenge. The best method is bold.}
\vspace{1pt}
\label{tab:accuracy}
\centering
\resizebox{0.97\linewidth}{!}{%
\begin{tabular}{cl|ccc|c}
\textbf{Setting}                  & \textbf{Method}                             & \textbf{ROD}          & \textbf{LineMOD}           & \textbf{NYU}   & \textbf{$\text{AvgA}_\%$}      \\ \hline
\multirow{6}{*}{\textbf{RGB}}     & FT                                          & 90.8                  & \textbf{96.7}              & 67.5           & 85.0               \\
                                  & FE                                          & 87.6                  & 13.9                       & 57.9           & 53.1               \\
                                  & PB \cite{piggyback}                         & 89.4                  & 95.9                       & \textbf{68.4}  & 84.6               \\
                                  & BAT \cite{quantized}                        & \textbf{92.9}         & 95.0                       & 68.1           & \textbf{85.3}      \\
                                  & RS  \cite{rebuffi_1}                        & 90.9                  & 89.7                       & 67.1           & 82.6               \\
                                  & RP \cite{rebuffi_2}                         & 91.3                  & 89.9                       & 68.3           & 83.2               \\ \hline
\multirow{6}{*}{\textbf{Depth}}       & FT                                          & 83.3                  & \textbf{87.0}              & 59.2           & \textbf{76.5}      \\
                                  & FE                                          & 78.3                  & 7.6                        & 43.1           & 43.0               \\
                                  & PB \cite{piggyback}                         & 83.5                  & 83.2                       & 56.6           & 74.4               \\
                                  & BAT \cite{quantized}                        & \textbf{83.7}         & \textbf{87.0}              & 57.3           & 76.0               \\
                                  & RS  \cite{rebuffi_1}                        & 83.6                  & 79.8                       & \textbf{61.2}  & 74.8               \\
                                  & RP \cite{rebuffi_2}                         & \textbf{83.7}         & 76.9                       & 57.3           & 72.6               \\ \hline
\multirow{6}{*}{\textbf{RGB-D}}   & FT                                          & \textbf{93.9}         & \textbf{96.9}              & 68.4           & \textbf{86.4}      \\
                                  & FE                                          & 91.6                  & 14.8                       & 60.2           & 55.5               \\
                                  & PB \cite{piggyback}                         & 89.6                  & 96.6                       & 66.7           & 84.2               \\
                                  & BAT \cite{quantized}                        & 93.5                  & 95.2                       & 68.1           & 85.6               \\
                                  & RS  \cite{rebuffi_1}                        & 93.7                  & 93.8                       & \textbf{70.9}  & 86.1               \\
                                  & RP \cite{rebuffi_2}                         & 93.8                  & 92.1                       & 67.5           & 84.5               \\\hline
\end{tabular}}
\squeezeup
\end{table}

Table \ref{tab:score} provides a better comparison of all the methods considering different evaluation metrics, settings and tasks. In the table, the \textit{Par} column reports the average per task of the ratio between the memory required by task-specific parameters $\rho_t $ and by parameters of the backbone network $ \rho_0 $, in formulas $\frac{1}{N}\sum_{t=1}^{N}\rho_t/\rho_0 $. 
For instance, FT corresponds to the value $1$ since it implies learning all the weights of a ResNet-18 for all the three tasks and FE corresponds to $0$ as only the classifier is learned.

It is interesting to discuss the different metrics. We focus on the comparison between BAT and RS in the RGB-D setting, which are the best performing methods among SMTL techniques. 
In term of the average accuracy AvgA$_\%$, RS considerably outperforms BAT. However, looking at Table \ref{tab:accuracy} we can see that RS is not always the best performer. Even if it achieves a very high accuracy in NYU, BAT outperforms RS in the LineMOD dataset and they are comparable in ROD. This result underlines the need for a metric that jointly consider all tasks. 
Indeed, the two methods have similar performance considering the Decathlon Score metric (DS) because DS 
ranks higher methods that are accurate in all the tasks. However, DS does not penalize methods which add several parameters. 
Oppositely, the Revised Decathlon Score (RevDS) strongly penalizes the addition of parameters. Thus, under RevD, we note a remarkable change in the ranking and BAT significantly outperforms RS. Actually, RS obtain a worse score than all the SMTL methods. 
Finally, considering the proposed Locally Linear score (LL), RS and BAT are comparable, as this metric is the only one which reflects the best trade-off between accuracy and use of extra parameters. 

\begin{table}[t]
\caption{Baselines score for the \challenge. The best method is bold, the second best is underlined.}
\label{tab:score}
\centering
\resizebox{1.\linewidth}{!}{%
\begin{tabular}{cl|c|ccc|c}
\textbf{Setting}                & \textbf{Method}                   & \textbf{Par} & \textbf{$\text{AvgA}_\%$} & \textbf{DS} & \textbf{RevDS} & \textbf{LL} \\ \hline
\multirow{6}{*}{\textbf{RGB}}   & FT                                & 1.00         & {\ul 85.0}    & \textbf{750}  & 8             & 0               \\ 
                                & FE                                & 0.00         & 53.1          & 229           & 229           & 0               \\ 
                                & PB \cite{piggyback}               & 0.03         & 84.6          & 577           & {\ul 463}     & 853             \\ 
                                & BAT \cite{quantized}              & 0.03         & \textbf{85.3} & {\ul 693}     & \textbf{556}  & \textbf{1196}   \\ 
                                & RS  \cite{rebuffi_1}              & 0.16         & 82.6          & 500           & 169           & 818             \\ 
                                & RP \cite{rebuffi_2}               & 0.13         & 83.2          & 542           & 228           & {\ul 922}       \\ \hline
\multirow{6}{*}{\textbf{Depth}}     & FT                                & 1.00         & \textbf{76.5} & \textbf{750}  & 8             & 0               \\
                                & FE                                & 0.00         & 43.0          & 213           & 213           & 0               \\
                                & PB \cite{piggyback}               & 0.03         & 74.4          & 598           & {\ul 480}     & {\ul 909}       \\
                                & BAT \cite{quantized}              & 0.03         & {\ul 76.0}    & {\ul 741}     & \textbf{595}  & \textbf{957}    \\
                                & RS  \cite{rebuffi_1}              & 0.16         & 74.8          & 582           & 197           & 865             \\
                                & RP \cite{rebuffi_2}               & 0.13         & 72.6          & 501           & 211           & 824             \\ \hline
\multirow{6}{*}{\textbf{RGB-D}} & FT                                & 1.00         & \textbf{86.4} & \textbf{750}  & 8             & 0               \\
                                & FE                                & 0.00         & 55.5          & 237           & 237           & 0               \\
                                & PB \cite{piggyback}               & 0.03         & 84.2          & 451           & {\ul 362}     & 560             \\
                                & BAT \cite{quantized}              & 0.03         & 85.6          & 514           & \textbf{413}  & {\ul 890}       \\
                                & RS  \cite{rebuffi_1}              & 0.16         & {\ul 86.1}    & {\ul 524}     & 177           & \textbf{891}    \\
                                & RP \cite{rebuffi_2}               & 0.13         & 84.5          & 481           & 203           & 818             \\ \hline
\end{tabular}%
}\squeezeup
\end{table}

\section{CONCLUSIONS}
We presented a novel benchmark for robot visual systems, namely the \challenge. This dataset allows to compare model in the sequential multi-task learning scenario where, starting from a pretrained deep architecture, the goal is to learn models for diverse visual tasks while i) obtaining performances as close as possible to fine-tuned task-specific architectures; ii) keeping as low as possible the number of additional parameters required per-task; and iii) not changing the performances on old tasks. We introduce a novel metric for the SMTL problem which takes into accounts both the performances and the number of parameters. We then evaluated state of the art SMTL methods on the new benchmark and released our source code for promoting further research on the topic. 
We hope that the \challenge \ will help to stimulate research in robot perception. 
While in the \challenge \ we considered a specific setting, our definition of tasks, metrics and baselines is modular and can be easily extended in other applications. 

\bibliography{bib}

\end{document}